\documentclass[journal]{IEEEtran}
\usepackage{amsthm,amssymb}
\usepackage{amsmath}
\usepackage{easyReview}
\usepackage{graphicx}
\usepackage{amsmath}
\usepackage{amsfonts}
\usepackage{algorithm}
\usepackage{algorithmic}
\usepackage{makecell}
\usepackage{booktabs}
\setcellgapes{3pt}
\usepackage{multirow}
\setcellgapes{3pt}

\usepackage[justification=centering]{caption}
\usepackage{color}
\usepackage{verbatim}
\usepackage{amssymb}

\usepackage{cite}

\ifCLASSINFOpdf
\else
\fi
\begin{document}
\bstctlcite{IEEEexample:BSTcontrol}
%
\title{LAMBO: Large AI Model Empowered Edge Intelligence}

\author{Li Dong, Feibo Jiang, \textit{Senior Member, IEEE}, Yubo Peng,  Kezhi Wang, \textit{Senior Member, IEEE}, Kun Yang, \textit{Fellow, IEEE}, Cunhua Pan, \textit{Senior Member, IEEE}, Robert Schober, \textit{Fellow, IEEE}}
\markboth{Submitted for Review}%
{Shell \MakeLowercase{\textit{et al.}}: Bare Demo of IEEEtran.cls for IEEE Journals}
%



\maketitle


\begin{abstract}
Next-generation edge intelligence is anticipated to benefit various applications via offloading techniques. However, traditional offloading architectures face several issues, including heterogeneous constraints, partial perception, uncertain generalization, and lack of tractability. 
In this paper, we propose a Large AI Model-Based Offloading (LAMBO) framework with over one billion parameters for solving these problems. We first use input embedding (IE) to achieve normalized feature representation with heterogeneous constraints and task prompts. Then, we introduce a novel asymmetric encoder-decoder (AED) as the decision-making model, which is an improved transformer architecture consisting of a deep encoder and a shallow decoder for global perception and decision. Next, actor-critic learning (ACL) is used to pre-train the AED for different optimization tasks under corresponding prompts, enhancing the AED's generalization in multi-task scenarios. Finally, we propose an active learning from expert feedback (ALEF) method to fine-tune the decoder of the AED for tracking changes in dynamic environments.
Our simulation results validate the advantages of the proposed LAMBO framework.

\end{abstract}


\IEEEpeerreviewmaketitle

\section{Introduction}
Multi-access edge computing (MEC) has been applied to allow for efficient and low-latency offloading of computational tasks from mobile devices via wireless networks \cite{8663993}. Integrating artificial intelligence (AI), e.g., deep learning with edge computing gives rise to the emerging concept known as edge intelligence\cite{ding2022roadmap}.

Recently, large AI models (LAMs) have been proposed as a novel intelligence paradigm that enables AI models with billions of parameters to achieve unprecedented general intelligence.
Pushing the adaptive learning and decision-making capabilities of LAMs to the network edge enables the provision of powerful personalized intelligent services to users while safeguarding data privacy. It is believed that the integration of LAMs and edge computing can form the foundation for the next generation of edge intelligence from the following aspects: LAM for edge and edge for LAM.

On the one hand, LAM for edge describes using LAMs to provide more data and intelligent solutions for edge computing, including managing and scheduling edge resources and allocating and coordinating edge tasks. This approach improves the efficiency, performance, reliability, and scalability of edge intelligence systems. Traditional optimization solutions for edge computing face challenges such as the dynamism and uncertainty of the edge environment; the diversity and distribution of edge systems; and the difficulty and complexity of edge optimization problems. 
LAMs can be extensively pre-trained in a broader range of datasets, exhibiting more robust generalization capabilities for dynamic and diverse tasks. Furthermore, with its billions of parameters, LAM surpasses traditional deep learning approaches in logical analysis and decision-making for complex optimization problems\cite{singhal2023large}.

On the other hand, edge for LAM focuses on the training and execution of LAMs on the edge. This approach enhances data privacy and security, reduces latency and communication overhead in data transmission, and provides personalized application scenarios and services. Traditional training and execution of AI models on edge systems face challenges such as the instability and heterogeneity of edge devices, as well as the unfairness and non-uniform distribution of edge data. LAMs can efficiently learn personalized data at the edge through fine-tuning with few parameters, thereby reducing the requirements for computing resources. Additionally, LAMs can possess powerful data representation and fault tolerance capabilities, reducing decision errors and biases caused by low-quality and non-uniform edge data \cite{tian2022fedbert}.

Against the above background, in this paper, we focus on how LAMs can empower edge intelligence from the perspective of offloading systems.
We first introduce the traditional deep offloading architecture and its main challenges. Subsequently, we describe key LAM technologies and the advantages of applying them to edge intelligence systems. Next, we propose a LAMBO framework with around 1 billion parameters for achieving high-quality task offloading and resource scheduling, targeting diverse tasks. Moreover, we use ``prompts" in LAMs to effectively manage and control LAMBO, enabling it to accomplish precise tasks through natural language instructions. 
Specifically, the prompt is deliberately assigned as two instructions (``Minimum latency" and ``Minimum energy") for different tasks. The contributions of this paper are summarized as follows:
{\color{black}
\begin{itemize}
\item \emph{High-quality Representation}: We introduce an input embedding (IE) approach that aims to convert inputs of various types (such as state information of the offloading system with heterogeneous constraints and instructions for diverse tasks) into cohesive feature embeddings while ensuring their normalization.

\item \emph{Excellent Performance}: We develop an asymmetric encoder-decoder (AED) architecture comprised of a deep encoder and a shallow decoder for modelling the decision-making process and producing excellent offloading decisions from the global perspective.

\item \emph{Superior Generalization}: We utilize actor-critic learning (ACL) to pre-train AED. Specifically, we generate a large number of unlabeled instances for various optimization tasks, which are then inputted into AED for extensive training to enhance generalization across multiple tasks.

\item \emph{Efficient Tractability}: We leverage active learning based on expert feedback (ALEF) to fine-tune the decoder of the AED while keeping the encoder frozen. This approach facilitates the efficient tracking of environmental variations. 
\end{itemize}
}

The remainder of this paper is organized as follows. The traditional deep offloading architecture and its challenges are presented in Section II. The key technologies and advantages of LAMs for MEC systems are described in Section III. The detailed design of the proposed LAMBO framework is provided in Section IV. Simulation results are presented in Section V, followed by a discussion of open issues in Section VI, and conclusions are drawn in Section VII.

\section{Traditional Deep Offloading Architecture and Challenges}

\subsection{Traditional Deep Offloading Architecture}
Traditional optimization algorithms for offloading decision-making and resource allocation may need to cope with resolving highly complex mixed-integer-nonlinear-programming (MINLP) problems via exhaustive search, branch-and-bound, or convex optimization-based techniques, which may have high complexity and low efficiency, that may not be suitable for dynamic environments. 

Deep offloading architectures have been proposed for intelligent offloading decision-making processes. These architectures apply deep learning methods (e.g., CNNs\cite{9167249}, RNNs\cite{8713801} and DNNs\cite{9275621}) with reinforcement learning \cite{8771176} or supervised learning \cite{9275621} to extract useful information and learn knowledge from MEC systems, and solve the offloading problem autonomously without human intervention. 
The traditional offloading architecture for MEC systems comprises three modules, as shown in the uppermost part of Fig. \ref{fig:fig1}, and described as follows \cite{9167249,8713801,9275621,8771176,8736011}:
\begin{figure*}[htpb]
	\centering
	\includegraphics[width=17cm]{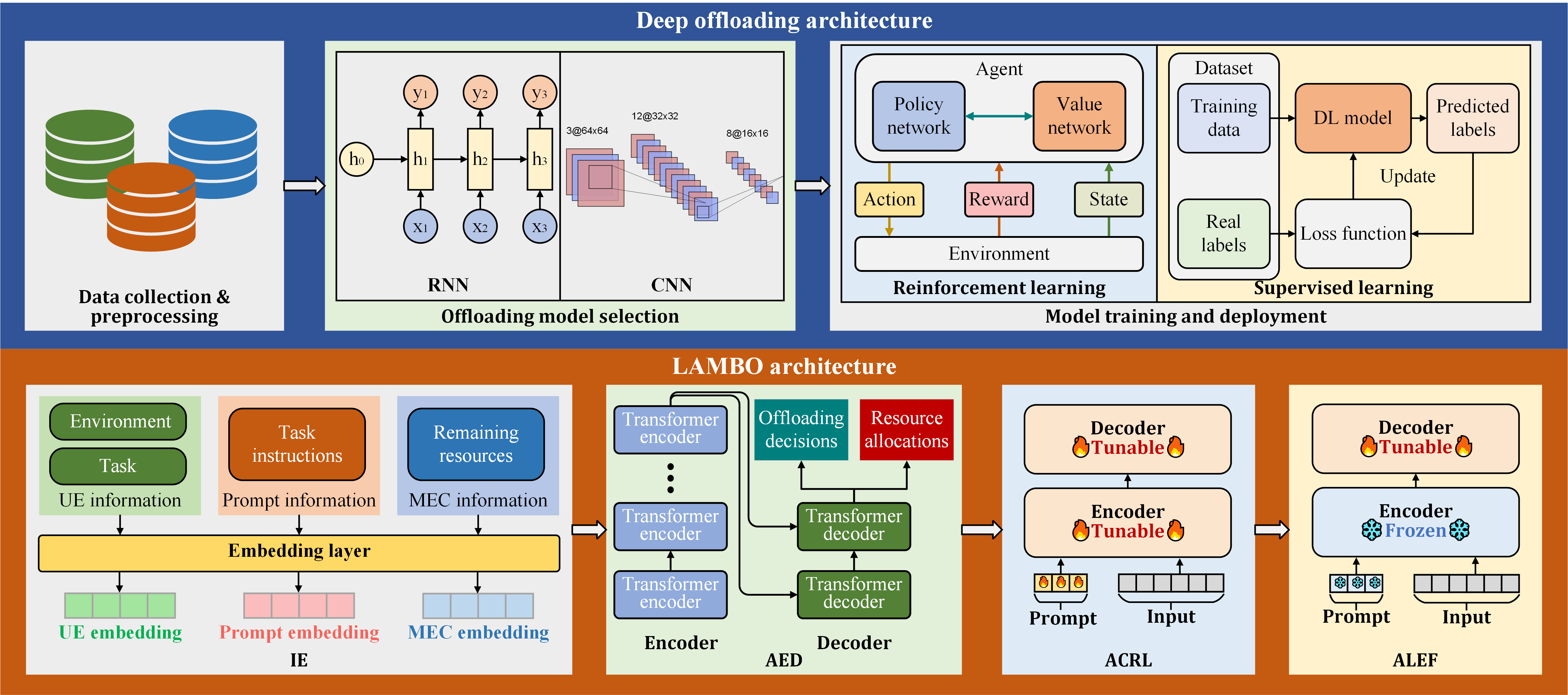}
	\caption{Deep offloading architecture versus LAMBO architecture.}
	\label{fig:fig1}
\end{figure*}

\subsubsection{Data Collection and Preprocessing}
Data collection is crucial for traditional offloading models that use deep learning in edge intelligence systems. Environmental data, such as sensory, task, and network information, can be gathered continuously as training data. A data preprocessing solution is necessary to discard or refine abnormal, incomplete, or duplicate data. Moreover, when the amount of training data is insufficient, data augmentation methods can be applied to generate more data for the offloading models.

\subsubsection{Offloading Model Selection}
CNNs and RNNs are common neural network architectures used in MEC systems for various tasks. CNNs are often used for grid-like data structures, such as path programming and offloading decision-making for mobile edge nodes. In contrast, RNNs are commonly designed to address time series prediction tasks such as offloading decision-making in dynamic environments.

\subsubsection{Model Training and Deployment}
Depending on the specific task in edge intelligence systems, an appropriate learning method can be chosen for the offloading model. Supervised learning can be employed when sufficient labelled data is available. However, when the data distribution, scale, quality, and privacy of the task are variable or unknown, reinforcement learning becomes the preferred training method. This approach utilizes observations and feedback from the environment to continually update the model, facilitating real-time learning and decision-making.


\vspace*{-2mm} 
\subsection{Research Challenges}
Although the traditional deep offloading architectures can solve the task offloading problem in edge intelligence systems, there are several challenges:

\subsubsection{Heterogeneous Constraints}
Edge systems are designed to accommodate highly heterogeneous environments with diverse requirements and constraints. Moreover, different edge nodes may possess varying computing and communication resources, resulting in distinct requirements for offloading decision-making and resource allocation. However, traditional deep offloading architectures are end-to-end learning systems and do not have dedicated components to handle these different requirements and constraints.

\subsubsection{Partial Perception}
Offloading decisions often rely on extensive sensor data, leading to high-dimensional features. The quality of communication channels between sensors and edge nodes can also influence user association with the edge server, which results in ample feature space regarding offloading decisions. CNNs and RNNs struggle with long-distance dependencies between features, causing partial perception issues that may lead to locally optimal offloading decisions.

\subsubsection{Uncertain Generalization}
Generalization implies deep learning models have consistent inference performance across different tasks. Specific offloading tasks may have critical optimization objectives (e.g., time-sensitive or energy-sensitive) and require rapid responses, placing stringent requirements on training time. This may challenge current deep offloading architectures with only one optimization goal, highlighting the need for superior generalization for different offloading tasks without retraining.

\subsubsection{Lack of Tractability}
Edge intelligence systems may be deployed in dynamic environments. It is difficult to ensure that the training process is general enough to capture the entire distribution of the input features as encountered in real-world scenarios, especially when the feature space is dynamic. Therefore, offloading models have to adaptively learn from dynamic scenarios and track the time-varying data in edge intelligence systems (e.g., network topologies and channel gains). While reinforcement learning is frequently employed for online decision-making in dynamic environments, it suffers from low learning efficiency and catastrophic forgetting in large-scale dynamic environments and multi-task dynamic environments.

\section{Key Technologies and Advantages of LAMs for Edge Intelligence Systems}

\subsection{Key Technologies}
In this subsection, we summarize several critical technologies of LAMs for their applications in edge intelligence systems.

\subsubsection{Prompt and Embedding}
Prompts and embedding are key techniques in LAMs used to perform preprocessing tasks for external instructions and information. They play a crucial role in integrating and processing these external inputs effectively in the LAM as follows \cite{naseem2020transformer}: 

\begin{itemize}
	\item \emph{Token Embeddings}: Token embeddings represent individual tokens, such as words or patches, in the input text. Each token is represented by a learned embedding that captures the lexical meaning of the word or patches. In MEC systems, environmental information (e.g., channel state information (CSI) for each user equipment (UE)) and task information (e.g., required data size and computing resources for each task) can be encoded as token embeddings for each UE.
	
	 \item \emph{Prompts}: Prompts are instructions or guides provided to the LAMs to execute specific tasks or generate desired outputs based on natural language. In MEC systems, prompts can serve as instructions to distinguish between different optimization tasks, such as minimizing energy consumption or system latency. 	For instance, when instructed to minimize latency as input, the LAMs can comprehend the semantic meaning of the instruction and generate task offloading and resource scheduling results that align with the goal of minimizing latency.
	 
\end{itemize}

\subsubsection{Transformer Model}
The transformer is a novel deep learning architecture proposed by Google \cite{vaswani2017attention}. Its key innovation is the attention mechanism, which enables the model to capture global dependencies between inputs in a sequence more effectively than traditional methods. 
Here is the structure  of the  transformer and its potential applications in MEC systems:

\begin{itemize}
\item \emph{Encoder-Decoder Models}:
Encoder-decoder models consist of separate encoder and decoder components in the transformer. The encoder processes the input sequence and generates contextualized representations for each token. The decoder then takes these representations and produces the output sequence. In MEC systems, the environment and task information of all UEs contribute to the input, and the encoder can effectively capture the system information in input embeddings, while the decoder generates accurate offloading decisions sequentially.


\end{itemize}

\subsubsection{Two-stage Training}
LAMs use supervised or unsupervised learning to pre-train the transformer on vast amounts of text data, and then they fine-tune the pre-trained transformer on specific downstream tasks. The technical details and potential applications of the two-stage Training are outlined below:

\begin{itemize}
\item \emph{Pre-training}:
Pre-training serves as the initial training stage of LAMs, which normally employs supervised or self-supervised learning techniques, enabling the model to capture statistical patterns, relationships, and contextual dependencies from extensive data \cite{zoph2020rethinking}. 
However, in MEC systems, 
collecting large amounts of supervised data could be expensive. Therefore, alternative pre-training methods that go beyond supervised and self-supervised learning could be considered for MEC systems.

\item \emph{Fine-tuning}:
Fine-tuning is the second training stage of LAMs, focusing on adjusting the parameters of the pre-trained LAM for a specific dataset or task \cite{wang2023fine}. This approach aims to enhance the model's generalization abilities while controlling its behaviour according to user preferences or task requirements. During the fine-tuning, the LAM is refined using a smaller dataset that is specifically relevant to the target task. In MEC systems, we can use fine-tuning to guide models in tracking the continuously changing communication environment.

\end{itemize}

\subsection{Advantages of Applying LAMs in Edge Intelligence Systems}
Based on the key technologies of LAMs presented in the previous subsection, we can outline the following benefits of applying LAMs in offloading systems:

\subsubsection{High-quality Representation of Heterogeneous Constraints}
Effective offloading and resource scheduling require precise modelling of the MEC system. However, different requirements and heterogeneous constraints of abundant edge nodes make this modelling difficult. LAMs have an additional embedding layer to handle the issue, allowing them to absorb all criteria and constraints, and then build extremely complex internal representations. This embedding layer is crucial for accurately modelling complex MEC systems. 

\subsubsection{Excellent Performance for Offloading Decision-making Process} 
LAMs are capable of autonomously conducting feature extraction and adaptive learning from MEC systems. Compared to small-scale models (e.g., CNNs and RNNs), LAMs often demonstrate superior inference accuracy. Through a multi-head self-attention mechanism in the transformer model, LAMs can establish different scales of dependencies between input information in an ample feature space. Hence, they can uncover more nuanced insights, which in turn contribute to excellent offloading decisions and resource allocation.

\subsubsection{Superior Generalization for Different Tasks}
By leveraging extensive datasets from various sources, LAMs acquire comprehensive knowledge across a broad spectrum of tasks, which can enhance their learning and generalization abilities, enabling effective modelling of diverse optimization objectives. Utilizing prompts during pre-training, LAMs can learn different optimization tasks and accurately perform specific functions without the need for retraining. This versatility makes LAMs highly adaptable and valuable for various optimization problems.

\subsubsection{Efficient Tractability in Dynamic Environments}
LAMs can efficiently utilize parallel computing resources, including distributed systems and specialized hardware accelerators such as graphics processing units (GPUs) or tensor processing units. It enables rapid fine-tuning for edge computing services. By employing adaptive fine-tuning, the pre-trained knowledge in LAMs can respond to dynamic environments, substantially boosting the robustness of offloading decisions and resource allocation in MEC systems.

\section{The LAMBO Framework}
This section introduces the LAMBO framework, including four components: IE, AED, ACL, and ALEF.

\subsection{Algorithm Overview}
As illustrated at the bottom of Fig. \ref{fig:fig1},
we first use IE to generate embeddings from the MEC system. 
Then, we propose a novel AED model to address the offloading and resource allocation. The AED model comprises a deep encoder and a shallow decoder. The encoder is employed to extract self-attention features for all UEs. These self-attention features, along with MEC and prompt embeddings, are fed into the decoder, which in turn generates optimal offloading decisions and resource allocation for each UE sequentially, ensuring they adhere to the constraints of the whole system. 

Moreover, we train the AED model in two stages: pre-training and fine-tuning. In the pre-training stage, we use an ACL to train the AED model. 
Then, in the fine-tuning stage, ALEF is used to keep track of environmental variations, detect unusual instances using maximum entropy, and label valuable instances based on expert feedback. Next, we provide a detailed description of each component. 



\subsection{Input Embedding} 
 We use IE to represent the input information to the edge intelligence system as embeddings in Fig. \ref{fig:AED}. 
 \begin{itemize}
 	\item \emph{UE embedding} denotes the environment (e.g., CSI) and task (e.g., computing and communication resource requirements) information of each UE.
 	\item \emph{Prompt embedding} denotes the instructions for different tasks with optimization goals, such as minimizing energy consumption or system latency.
 	\item \emph{MEC embedding} denotes the remaining resource constraints of edge servers, which are dynamically updated with each UE's resource allocation decision.
 \end{itemize} 


Firstly, we tokenize the textual prompt into a numerical representation. Then, we normalize all the information (UE, MEC, and prompt) and input them into IE, which is a learnable hidden layer that structurally enables the mapping of diverse input information into fixed-length vectors (embeddings) for the attention model.

\subsection{Asymmetric Encoder-Decoder}
\begin{figure}[htbp]
	\centering
	\includegraphics[width=9cm]{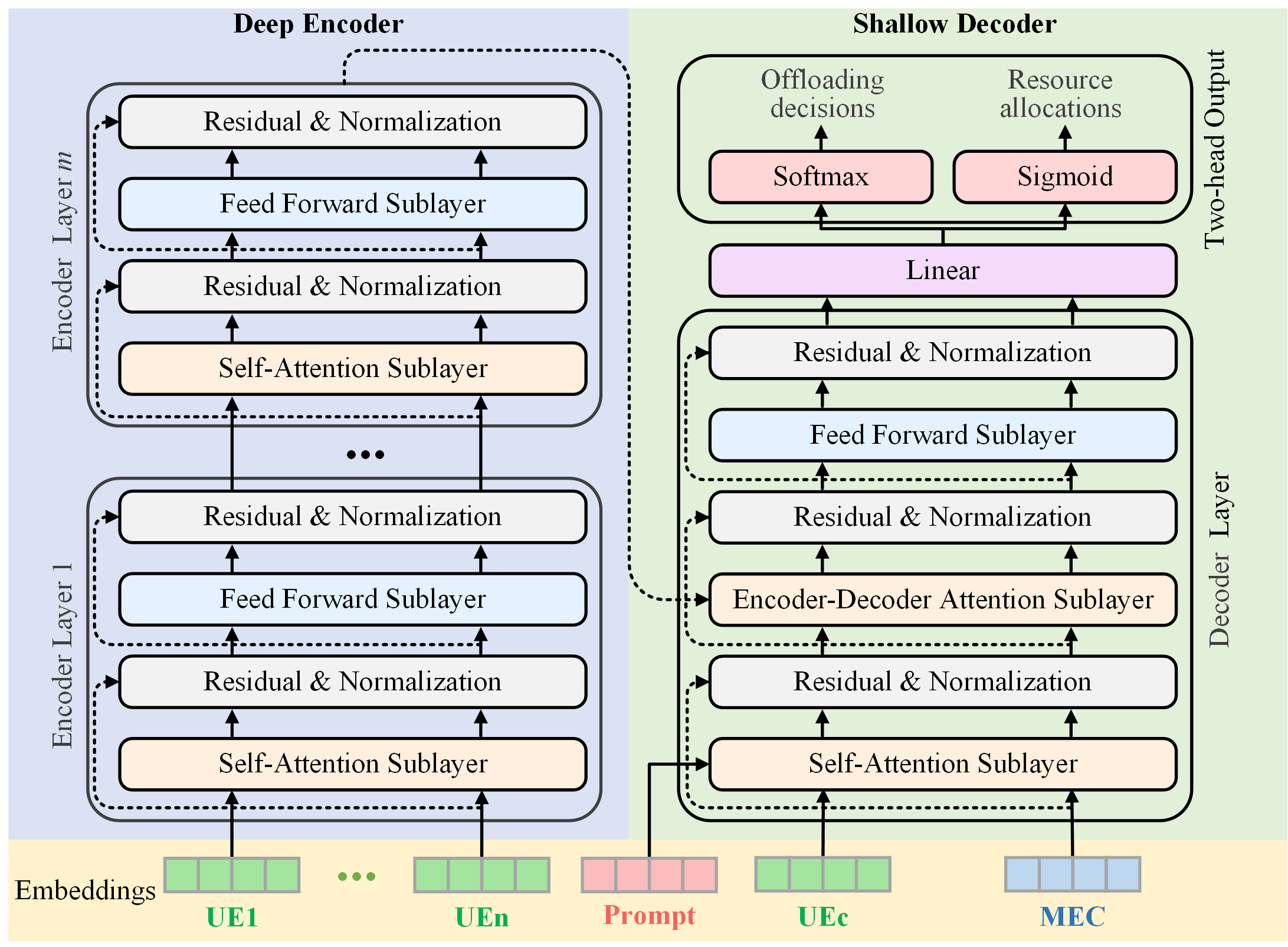}
	\caption{Structures of IE and AED.}
	\label{fig:AED}
\end{figure}

The AED model comprises a deep encoder with multiple encoder layers, enabling effective learning from input embeddings. Furthermore, it incorporates a shallow decoder with a single decoder layer, allowing for sequential generation of offloading decisions and resource allocation for each UE. The structures of the encoder and the decoder in AED are illustrated in Fig. \ref{fig:AED}.

\subsubsection{Deep Encoder}
Each encoder layer typically has the following sub-components:
\begin{itemize}
	
\item \emph{Self-Attention Sublayer}: This sublayer allows the model to focus on different parts of the input sequence of embeddings simultaneously, capturing dependencies and relationships between embeddings.

\item \emph{Feed-Forward Sublayer}: This sublayer applies a point-wise feed-forward transformation to each embedding independently, allowing non-linear transformations to be applied to the input sequence.

\item \emph{Residual and Normalization Sublayer}: After each Self-Attention Sublayer and Feed-Forward Sublayer, a residual connection is applied and normalization is applied to normalize the activations across the feature dimension.

\end{itemize}

\subsubsection{Shallow Decoder}
The decoder layer typically has the following sub-components:
\begin{itemize}
\item \emph{Self-Attention Sublayer}: This sublayer follows a similar procedure to the one used in the encoder, and residual connection and layer normalization are also applied to all sublayers. 
\item \emph{Encoder-Decoder Attention Sublayer}: This sublayer attends to the encoder's output features, allowing the decoder to focus on relevant features from the input.
\item \emph{Linear Projection and Two-head Output}: Following the decoder's final layer, output representations are projected to the output size through a linear transformation. A two-head output layer is then applied to solve the optimization problem, e.g., MINLP efficiently, in which a softmax function is introduced to generate the offloading decisions, and a sigmoid function is implemented to create the allocated resources. 
\end{itemize}

The advantage of employing an asymmetric structure lies in the ability to adjust only the parameters of the shallow decoder during the fine-tuning phase, thereby facilitating rapid learning of LAMBO.



\subsection{Pre-training: Actor-Critic Learning}
\begin{figure}[htbp]
	\centering
	\includegraphics[width=8.5cm]{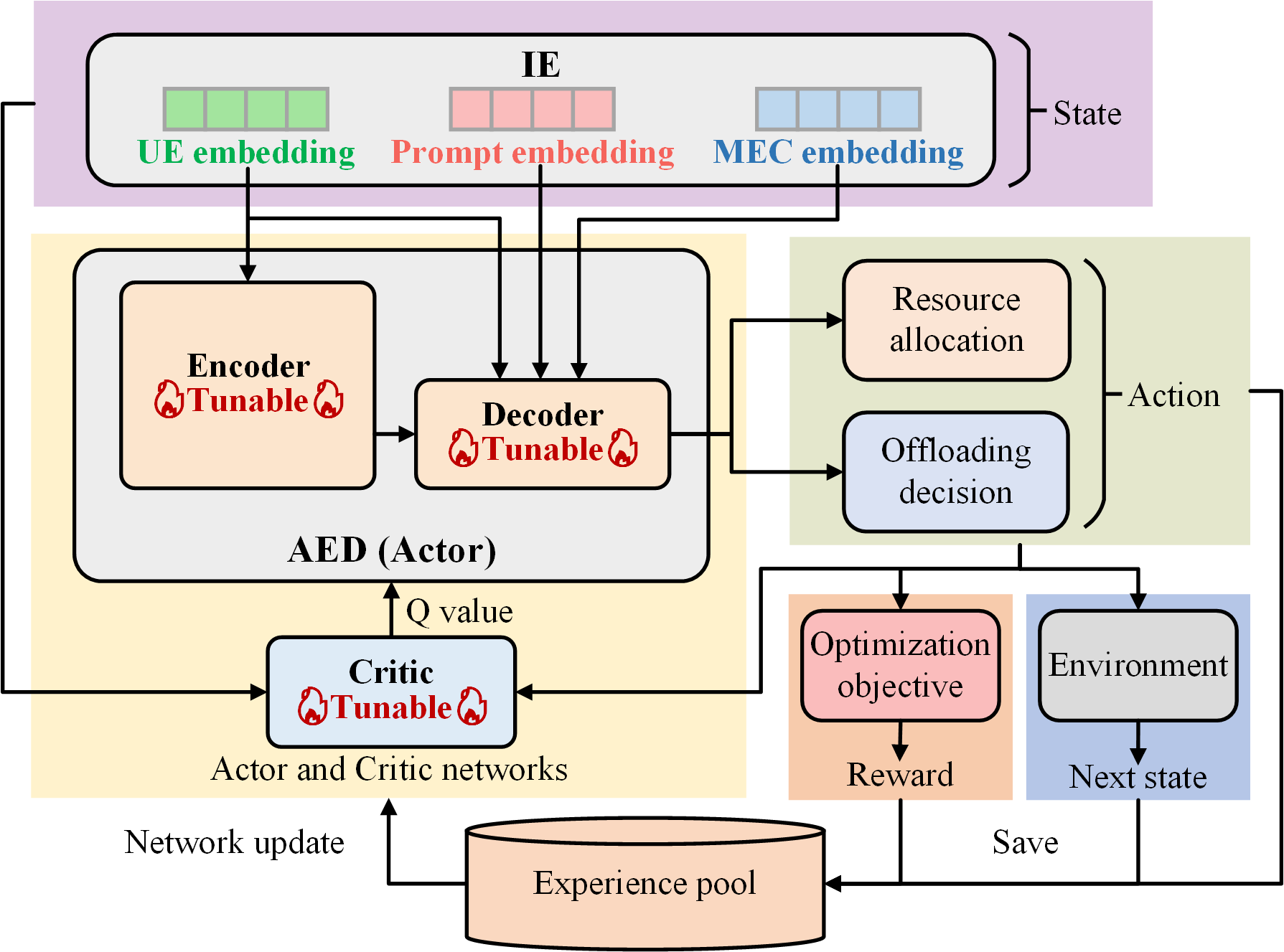}
	\caption{Workflow of ACL.}
	\label{fig:ACRL}
\end{figure}

We employ ACL in the first training stage to pre-train AED, which is detailed below and illustrated in Fig. \ref{fig:ACRL}:
\subsubsection{Actor Network Design}
In MEC systems, we designed an actor-network to learn and improve decisions for LAMBO. The actor-network utilizes the AED model, taking MEC embedding, UE embedding, and prompt embedding as inputs, and produces offloading choices and resource allocation as outputs. The actor-network is trained using the policy gradient method to boost its exploration.
\subsubsection{Critic Network Design}
We design an additional critic network to evaluate the reward of the actor network's policy, which corresponds to the optimization objective for the current prompt in MEC systems. The critic network adopts the same output structure as the actor-network to enhance the stability and accuracy of reward prediction.
\subsubsection{Training Iteration and Policy Update}
We set appropriate parameters, such as the learning rate, to facilitate the training process. Then, an iterative training process begins until the actor and critic networks converge.

\subsection{Fine-tuning: Active Learning based on Expert Feedback}
Since the network topology and CSI are time-varying, we introduce ALEF in the second fine-tuning stage to proactively track the variations of the environment with fewer labelled instances collected from the environment. The fine-tuning process is detailed below, and illustrated in Fig. \ref{fig:ALEF}:

\subsubsection{Online Offloading Decision-making Process}
The pre-trained AED model is deployed for the downstream task. In this step, the encoder parameters are frozen, and the lightweight decoder with a few parameters is utilized to make offloading decisions efficiently for unlabeled instances sampled from MEC systems operating in realistic scenarios. 

\subsubsection{Maximum Entropy Query}
A query strategy based on the offloading decision is introduced to identify the importance of instances in dynamic environments and select the cases that would yield the highest performance gain for the decoder. The instance with the maximum entropy is chosen since it reflects the most significant uncertainty of the offloading decision \cite{10.1145/3472291}.

\subsubsection{Expert Appointment and Decoder Update}
``Expert" is the specific algorithm for improving the selected instances. For instance, Mixed-integer programming solvers (e.g., CPLEX), heuristic algorithms (e.g., differential evolution (DE)), and policy optimization algorithms (e.g., policy gradient (PG), soft Actor-Critic).
Finally, the lightweight decoder is updated by the expert from the selected instances for tracking the varying environments. 

\vspace*{-2mm}
\section{Case Study}
\subsection{Problem Formulation}
We consider a standard MEC system consisting of several edge servers and UEs. Each UE has a single compute-intensive task and adheres to a binary offloading policy. 
Unlike traditional deep offloading architectures, the proposed architecture can make offloading decisions for different optimization tasks simultaneously based on prompts without retraining the model.
To verify the validity of prompts, we define two different optimization tasks for the considered MEC system, where the prompts and their meanings are as follows:

\begin{figure}[htbp]
	\centering
	\includegraphics[width=9cm]{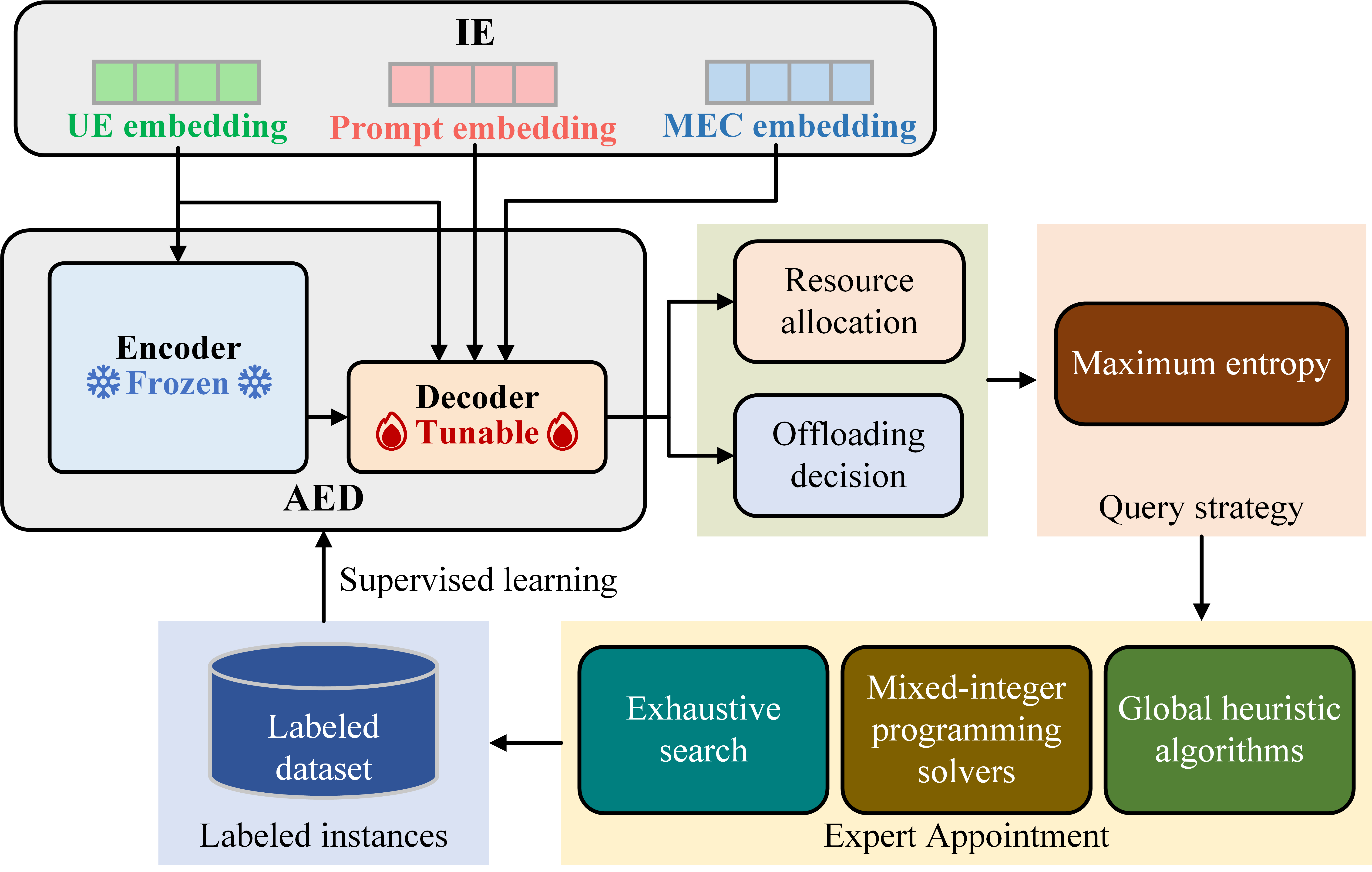}
	\caption{Workflow of ALEF.}
	\label{fig:ALEF}
\end{figure} 

\begin{itemize}	
	\item \emph{Minimum Latency}: User association and resource allocation are optimized by minimizing the average latency for each UE. Latency is composed of transmission latency and execution latency for the remote task or local execution latency \cite{6787113}.
	
	\item \emph{Minimum Energy}: User association and resource allocation are optimized for the minimization of the total energy consumption of all UEs. Energy consumption comprises transmission energy for the remote task and local execution energy \cite{6787113}.

\end{itemize}

The optimization problem must satisfy constraints: (1) the maximum latency constraint for each task; (2) the maximum resource availability in each edge server; (3) all tasks can be executed either locally at the UE or on a single edge server.

\subsection{Simulation Settings}

To evaluate the performance of the proposed framework, we consider a scenario involving 4 edge servers and 50 UEs in an 50 m $\times$ 50 m square area. 
All edge servers and UEs are randomly distributed in the area, and the 
UE's maximal velocity is 1.2 m/s.
The CSI, which already encapsulates the spatial relationship and channel fading relationship between the UE and edge server is calculated from our previous work \cite{9275621}.
The maximum latency for all UEs is set to 1.5 seconds. The computing resources of each edge server are allocated as follows: $1.5 \times {10^{10}}$, $1.5 \times {10^{10}}$, $3 \times {10^{10}}$, and $5 \times {10^{10}}$ cycles/s, respectively.
For each UE, the average data size and the corresponding required computing resources for tasks are $2 \times 10^{8}$ kB and $1 \times 10^{9}$ cycles/s, respectively.
Both transmitting power and local execution power are set to 1 W. 

\subsection{Simulation Contenders}
To showcase the benefits of the proposed LAMBO framework, we compare it with the following contenders:
\begin{itemize}	
	\item Local: All tasks are executed locally.
	\item Random: All tasks are executed on the UE or one MEC randomly. 
	\item DE: Offloading decisions are optimized by the DE algorithm.
	
	\item DROO: Offloading decisions are optimized by the traditional deep offloading architecture with reinforcement learning \cite{8771176}.
	The agent in DROO comprises 60 fully connected FC layers. The total number of parameters in the agent is 326,662.
	\item  ARE: Offloading decisions are optimized by the traditional deep offloading architecture with supervised learning \cite{9275621}.
	The DNN in ARE comprises 60 FC layers. 
The total number of parameters in the DNN model is 326,662.
	
	\item LAMBO\_M: A medium-scale AED model is utilized to optimize offloading decisions. In LAMBO\_M, the encoder comprises 60 encoder layers, and the decoder incorporates 6 decoder layers. 
The total number of parameters in LAMBO\_M is 579,920,902.

	\item LAMBO\_L: A large-scale AED model is utilized to optimize offloading decisions. In LAMBO\_L, the encoder consists of 120 encoder layers, and the decoder incorporates 12 decoder layers. 
	The total number of parameters in LAMBO\_L is 1,159,515,142.	
\end{itemize}
\begin{figure}[htbp]
	\centering
	\includegraphics[width=8.5cm]{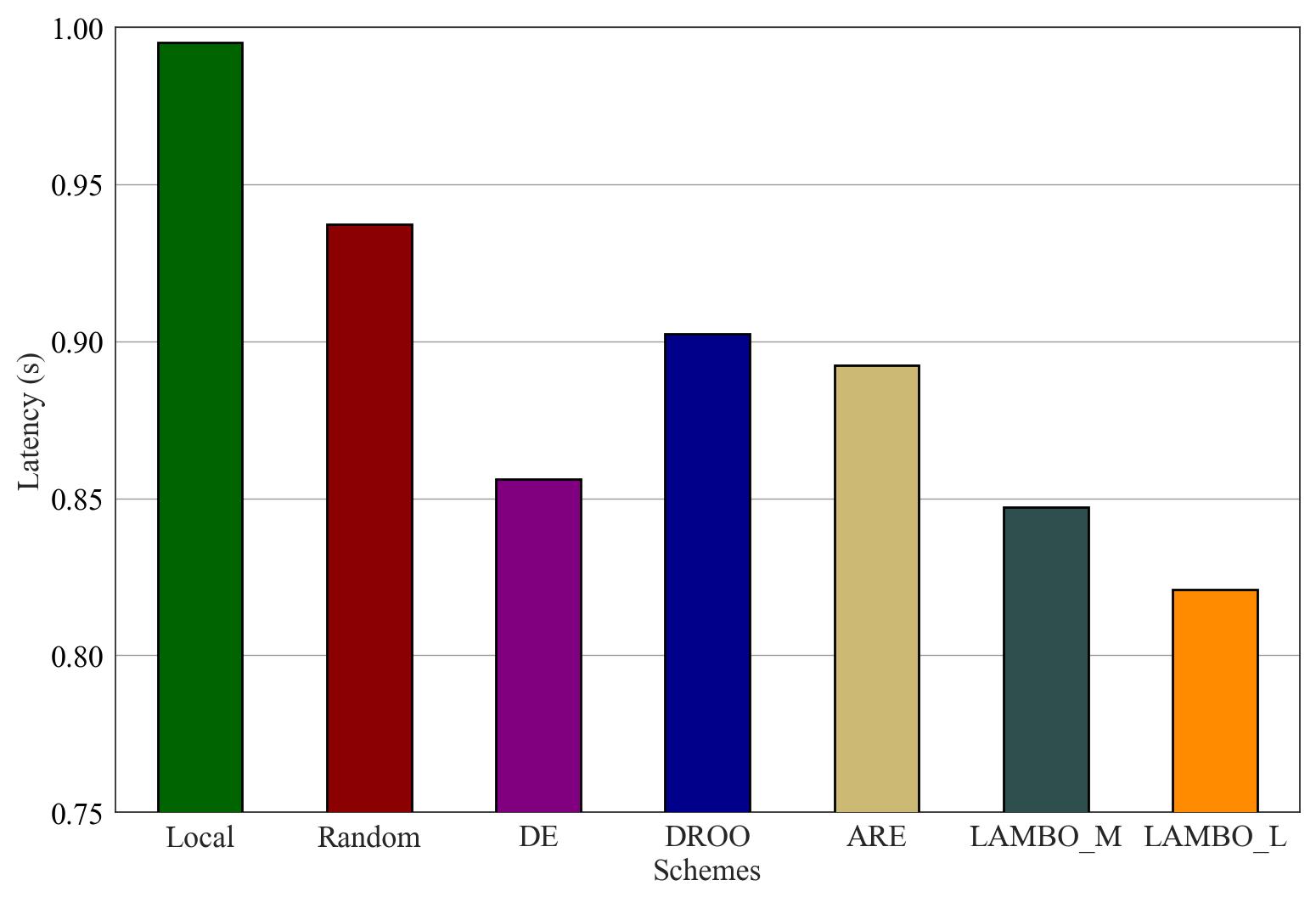}
	\caption{Comparison of different schemes in task latency.}
	\label{fig:exp1}
\end{figure} 
\vspace*{-3mm} 
 \subsection{Evaluation Results}
We evaluate the proposed LAMBO using two prompts. The simulation results of all contenders for the minimum latency prompt are presented in Fig. \ref{fig:exp1}, with the y-axis representing the average latency of each UE.
We can see that LAMBO\_L achieves the lowest latency, while DE performs third best, closely behind LAMBO\_M.

\begin{figure}[htbp]
		\centering
		\includegraphics[width=8.5cm]{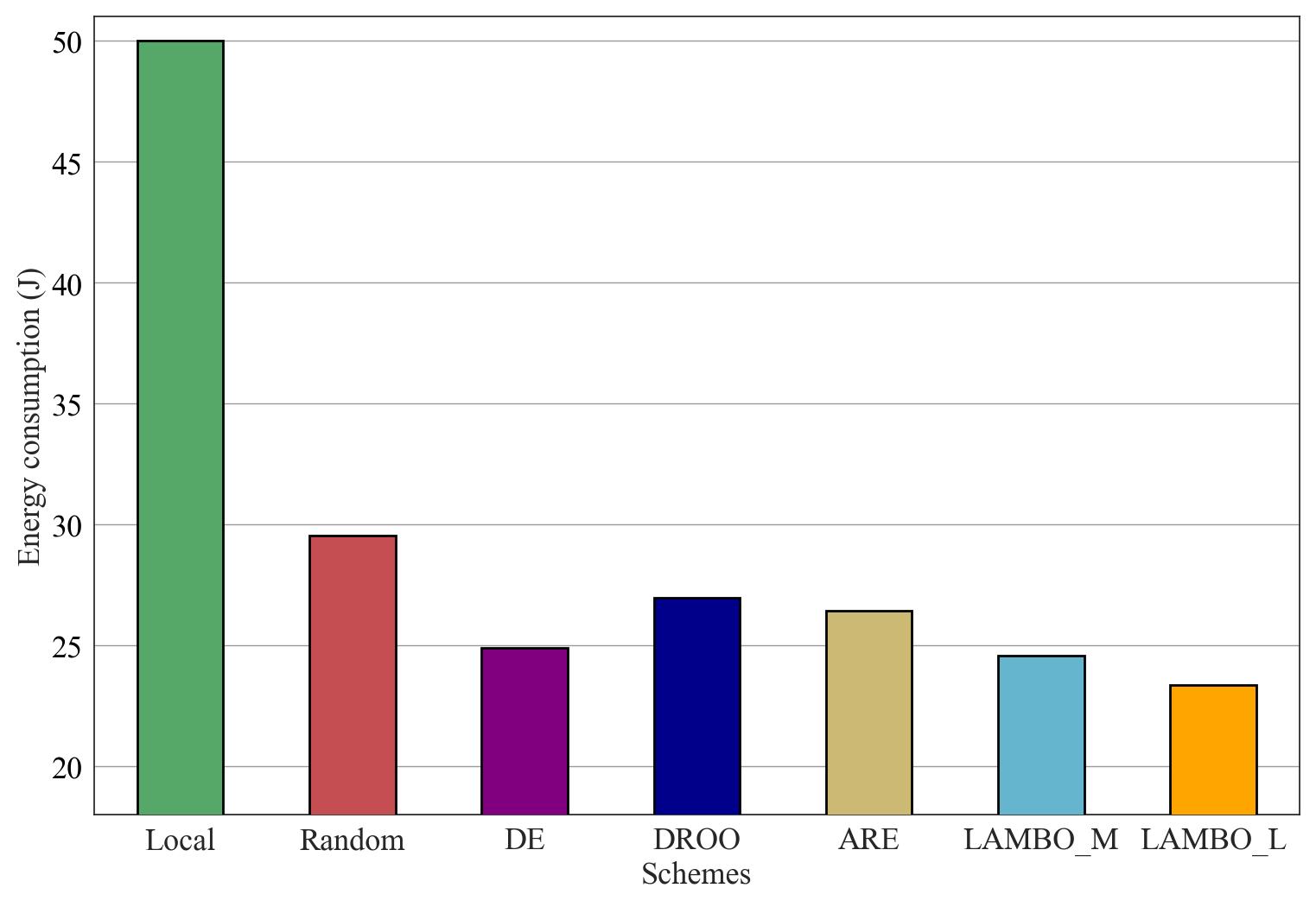}
		\caption{Comparison of different schemes in energy consumption.}
		\label{fig:exp2}
	\end{figure}

Similar observations can be made in Fig. \ref{fig:exp2} for the minimum energy prompt, where the y-axis represents the total energy consumption for all UEs. 
It can be inferred from the results that the AED exhibits a remarkable learning capacity, resulting in higher accuracy of offloading decisions compared to other schemes.
Furthermore, the learning capability improves on all prompts as the number of parameters increases, suggesting that LAMBOs have greater potential to address a variety of challenges in MEC systems compared to traditional offloading methods such as ARE and DROO.
Next, compared to LAMBO\_L, the performance of the LAMBO\_M model consistently lags behind. This could be attributed to LAMBO\_L having a higher number of parameters, thus possessing stronger learning and decision-making capabilities.
\vspace*{-2.5mm} 
\section{Open Issues and Future Directions}
\subsubsection{Limited Memory and Storage}
Edge servers and devices like base stations, cars, and IoT devices have much less RAM and flash storage compared to cloud centers. This restricts the maximum model size that can be deployed. LAMs easily exceed 50 GB, which cannot fit on some edge servers and devices. Further exploration of compression techniques such as sparse attention, quantization, and distillation is warranted in future research, as they provide valuable means to reduce the model size and minimize storage requirements.

\subsubsection{Constrained Computing Power} Edge servers and devices have relatively simple CPUs or microcontrollers compared to the powerful GPUs used to train LAMs. The computation needed for inference places a heavy load on edge processors. This leads to slow response times and high battery/power drain. Optimized inference engines such as multi-query attention and latent consistency tailored for edge devices are an important research direction for the future.
\subsubsection{Substantial Communication Overhead} 
Edge devices face bandwidth limitations, hindering their ability to handle high data throughput. This leads to transmission bottlenecks and inefficiencies for LAMs. Unstable connectivity in edge environments adds complexity, causing disruptions and data inconsistencies. Research on novel communication paradigms, such as semantic communication systems, holds the potential to provide innovative solutions for reducing the communication overhead of LAMs at the edge in the future.

\section{Conclusion}
In this paper, we have studied an LAM-based offloading framework for MEC systems. We first summarized the main challenges of the traditional deep offloading architecture, and then we described the key technologies of LAMs and the advantages of applying LAMs in MEC systems. Next, we proposed the LAMBO framework for MEC systems, in which IE was applied to represent the system information with constraints and prompts, AED was presented to model the offloading decision process, ACL was introduced to pre-train the AED for different tasks, and ALEF was used to fine-tune the decoder of AED for tracking the dynamic environments.
\bibliographystyle{IEEEtran}
\bibliography{bare_jrnl_bobo}

\section*{Biographies}
\textbf{Li Dong} (Dlj2017@hunnu.edu.cn) is currently an Associate Professor at Hunan University of Technology and Business, China.

\textbf{Feibo Jiang} (jiangfb@hunnu.edu.cn) is currently an Associate Professor at Hunan Normal University, China.

\textbf{Yubo Peng} (pengyubo@hunnu.edu.cn) is currently pursuing the master’s degree at Hunan Normal University, China. 

\textbf{Kezhi Wang} (Kezhi.Wang@brunel.ac.uk) is a Senior Lecturer with the Department of Computer Science, Brunel University London, U.K.

\textbf{Kun Yang} (kunyang@essex.ac.uk) is currently a Chair Professor in the School of Computer Science \& Electronic Engineering, University of Essex, U.K. 

\textbf{Cunhua Pan} (cpan@seu.edu.cn) is currently a full professor in Southeast University, China. 

\textbf{Robert Schober} (robert.schober@fau.de) is an Alexander von Humboldt Professor and the Institute for Digital Communications (IDC) in Friedrich-Alexander University (FAU) Erlangen-Nürnberg, Erlangen, Germany.

\end{document}